\DocumentMetadata{%
	pdfstandard=A-3b,
	pdfversion=1.7,
	lang=en-US,
}

\documentclass[11pt]{article}
\usepackage{amsmath}
\usepackage{amssymb}
\usepackage{booktabs} 
\usepackage{tikz}
\DeclareMathOperator*{\argmax}{arg\,max}

\begin{document}

\title{Flow Battery Manifold Design with Heterogeneous Inputs Through Generative Adversarial Neural Networks} 

\author{
Eric Seng\thanks{Department of Electrical and Computer Engineering, Duke University, Durham, North Carolina}
\and Hugh O'Connor\thanks{School of Chemistry and Chemical Engineering, Queen's University Belfast, Belfast, Northern Ireland}
\and Adam Boyce\thanks{School of Mechanical and Materials Engineering, University College Dublin, Dublin, Ireland}
\and Josh J. Bailey\thanks{School of Chemistry and Chemical Engineering, Queen's University Belfast, Belfast, Northern Ireland}
\and Anton van Beek\thanks{School of Mechanical and Materials Engineering, University College Dublin, Dublin, Ireland, \texttt{anton.vanbeek@ucd.ie}}
}
\maketitle

\begin{abstract}
Generative machine learning has emerged as a powerful tool for design representation and exploration. However, its application is often constrained by the need for large datasets of existing designs and the lack of interpretability about what features drive optimality. To address these challenges, we introduce a systematic framework for constructing training datasets tailored to generative models and demonstrate how these models can be leveraged for interpretable design. The novelty of this work is twofold: (i) we present a systematic framework for generating archetypes with internally homogeneous but mutually heterogeneous inputs that can be used to generate a training dataset, and (ii) we show how integrating generative models with Bayesian optimization can enhance the interpretability of the latent space of admissible designs. These findings are validated by using the framework to design a flow battery manifold, demonstrating that it effectively captures the space of feasible designs, including novel configurations while enabling efficient exploration. This work broadens the applicability of generative machine-learning models in system designs by enhancing quality and reliability.

\end{abstract}


\section{Introduction}
\begin{figure}
    \centering
    \includegraphics[width=0.99\linewidth]{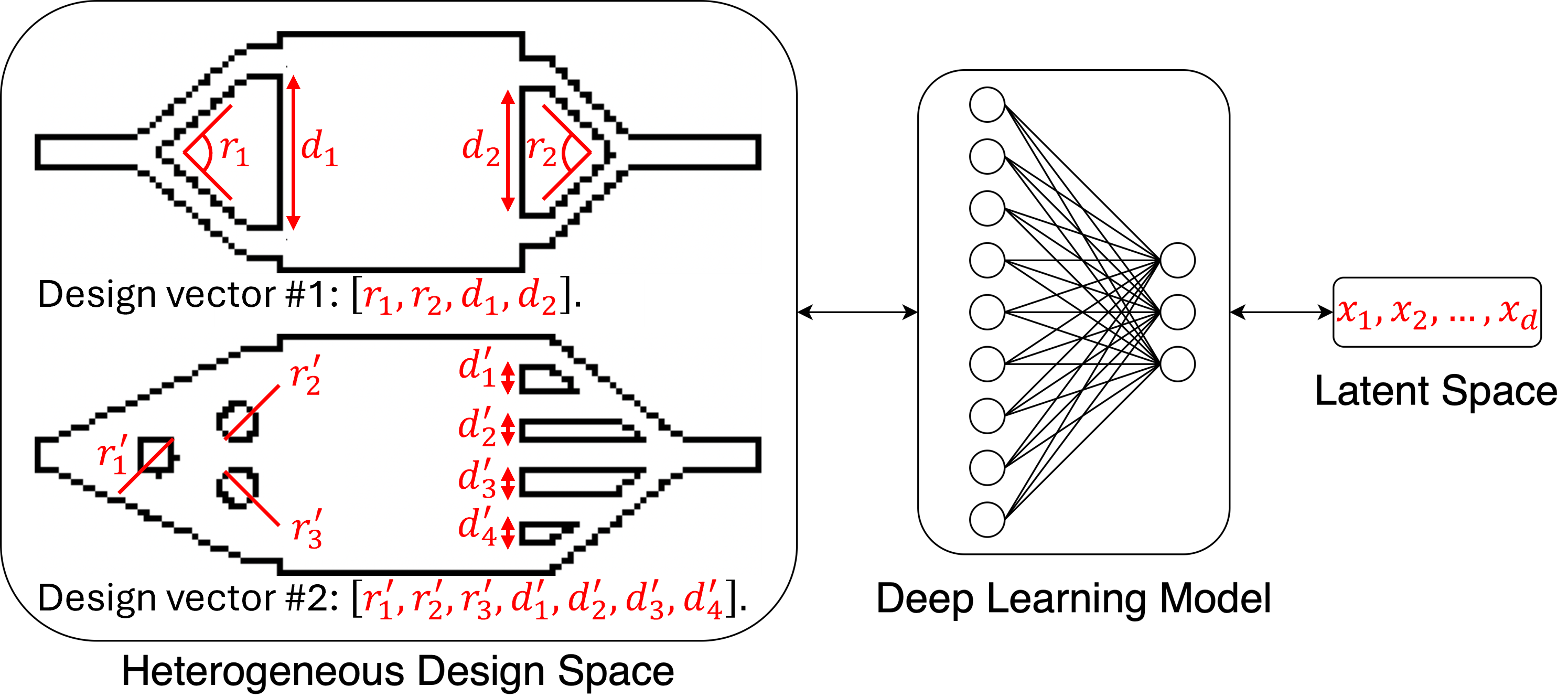}
    \caption{Graphical representation of the use of machine learning models to learn continuous latent space representations of heterogeneous design problems.}
    \label{fig:overarching}
\end{figure}

Generative machine-learning models have emerged as powerful tools for exploring complex and high-dimensional design spaces that are challenging to parametrize through a small set of independent design variables (e.g., the shapes of bicycles, \cite{regenwetter2022} and pipe systems \cite{urata2024}). Specifically, generative machine-learning models enable engineers to capture a system’s admissible design space through a set of latent variables. These latent variables typically follow a predefined distribution from which new designs can be sampled/generated. When combined with optimization techniques \cite{chen2020}, generative models can guide design decisions efficiently by focusing exploration on regions of the design space with high potential. However, a fundamental challenge in deploying these models lies in the requirement for extensive training datasets, which are unavailable for many design domains. This limitation restricts the application of generative machine-learning models to fields with abundant pre-existing data, leaving many innovative domains untapped. To address this gap, we introduce a systematic method for generating large datasets from heterogeneous design spaces (i.e., spaces characterized by the presence of plausible designs that cannot be represented by a small enough and uniform set of variables). For example, the two manifolds represented in the left panel of Figure~\ref{fig:overarching} have different numbers of inlets and classes of shapes that would result in different sets of design variables. By enabling the exploration of such diverse spaces, we aim to unlock new opportunities for data-driven design in fields where traditional datasets are sparse or non-existent.

Several methods have been proposed to support design optimization in heterogeneous design spaces, particularly within the subclass of mixed-variable optimization problems.  In this work, we use the term heterogeneous design space to refer to the scenario where different system instances are represented by design variables of varying lengths, each with distinct physical meanings, yet corresponding to the same artifact. In such problems, qualitatively different design alternatives—such as the two shown in Figure~\ref{fig:overarching}, can be represented using categorical variables \cite{Lucidi2005,Zhang2020}. These approaches are appropriate when the different design categories share the same set of continuous design variables. However, as is the case with these flow channels, this assumption does not always hold. For example, in interplanetary trajectory planning, the underlying design representations may vary significantly, making evolutionary optimization methods more suitable \cite{Hui2015,Gamot2023}. To mitigate the computational cost associated with these methods, techniques such as subspace optimization or variable-size design space kernels have been proposed \cite{Pelamatti2021}. However, these approaches still require that distinct designs share a subset of common variables, and they remain computationally intensive when applied to high-dimensional design spaces with strongly correlated inputs.

Heterogeneous design spaces offer significant potential for generative machine-learning models in design exploration and optimization owing to their compatibility with different types of data modalities. By representing these heterogeneous design spaces in a continuous latent space (i.e., the right-facing arrows in Figure~\ref{fig:overarching}), new designs can be efficiently generated and optimized with respect to one or multiple objectives (i.e., the left-facing arrows in Figure~\ref{fig:overarching}). For this purpose, various types of models have been proposed, including i) Unsupervised learning methods such as autoencoders \cite{griffiths2020}, variational autoencoders (VAEs) \cite{kingma2022}, generative adversarial networks (GANs), \cite{yuan2023}, and diffusion models \cite{ho2020}, ii) self-supervised learning such as info-GANs \cite{chen2020}, and iii) supervised learning such as conditional GANs \cite{matsuda2022}. Autoencoders and VAEs reduce the dimensionality of the design space by learning compact latent representations that preserve essential features of the input data, facilitating reconstruction and interpolation between designs. GANs, on the other hand, learn to generate realistic designs by pitting a generator against a discriminator in an adversarial training process, making them well-suited for exploring diverse design solutions \cite{goodfellow2020}. In addition, diffusion models can be used for dimensionality reduction by learning a low-dimensional latent representation through training a model to denoise high-dimensional noise into meaningful samples. Finally, conditional GANs and Info-GANs provide additional structure to the latent space, enabling the control of specific design attributes and enhancing interpretability, with Info-GANs \cite{chen2016}.  

A domain where generative machine-learning models are particularly relevant is for the design from images as this presents a high-dimensional space of correlated input variables. Such challenges are typically addressed through topology optimization methods that involve the optimization of material distribution to meet specific objectives \cite{wang2022}. However, such methods are often computationally intensive as a design’s performance needs to be evaluated/tested using simulation or physical experiments. To this end, generative models provide a profound potential as they can use the structure in a training data set of admissible designs to expedite design exploration (i.e., reduce the total number of experiments). These models leverage a design’s latent space representation to enable compatibility with optimization techniques such as evolutionary algorithms (e.g., particle swarm optimization and genetic algorithm) and Bayesian optimization \cite{vanbeek2021, vanbeek2024}. The latent space acts as a reduced-dimensional search space, allowing optimization algorithms to efficiently test promising designs \cite{giannone2023}. However, these approaches are fundamentally reliant on the availability of a training dataset composed of admissible designs or the ability to generate such data (e.g., solid isotropic material with the penalization method in topology optimization). \cite{bendsoe2013}.

In this paper, we introduce a systematic framework to generate training data sets of engineering drawings that can be used to train generative models and efficiently explore heterogeneous design spaces. This approach is conceptually similar to the variable size design space kernel approach in \cite{Pelamatti2021}, but computationally more efficient as the spatial relation between admissible designs is approximated only ones (i.e., before initiating the optimization procedure). In addition, we show how the integration of these generative machine-learning models within a Bayesian optimization framework can provide additional interpretability of the latent space of admissible designs.  Specifically, we introduce a three-step framework to establish a dataset of representative designs, use that dataset to train an Info-GAN and leverage the learned latent space representation in a Bayesian optimization loop to discover new designs with strong performance.  Through this framework, we enable engineers to efficiently explore heterogeneous design spaces through a manageable and interpretable number of design variables, reducing the necessary experimental cost. We demonstrate the performance of the presented framework through the design of a flow battery manifold, showcasing improved charge voltage and charge capacity over a manageable number of costly simulations that couple computational fluid dynamics and electrochemistry simulations. The framework generalizes to other problems that involve heterogeneous design space (e.g., building design and automotive engineering). 


The remainder of this paper is structured as follows. In Section 2 we will introduce generative machine-learning models and multi-objective Bayesian optimization that are foundational to the introduced framework. Next, in Section 3 we will introduce the three-phase design framework for design with heterogeneous design spaces. Subsequently, in Section 4 we will demonstrate the application of the framework for the design of flow batteries. Finally, in Section 5 we will discuss the findings, highlight key insights, and outline potential directions for future research.


\section{Background}
In this section, we will introduce the fundamental methods underpinning our approach, generative machine-learning models and multi-objective Bayesian optimization.

\subsection{Generative Machine Learning Models}
Generative machine-learning models provide a compelling tool to address many engineering challenges, due to their ability to learn the structure that underpins high-dimensional correlated datasets.  For example, generative machine-learning models have been applied for image generation, natural language processing, and data augmentation \cite{ramesh2022}.  Broadly, generative models are divided into two groups: unsupervised models, which learn the full design space without labels, and supervised (conditional) models, which use labels to generate data in explicit conditions \cite{Mirza2014}.  Focusing on the unsupervised generators, there are a few commonly used models variational autoencoders (VAE), GANs, and diffusion models \cite{regenwetter2022b}.  

Autoencoders are a type of unsupervised learning model that uses a neural network to reduce the dimensionality of complex design representations.  This is achieved through two networks, an encoder and a decoder.  The encoder is tasked with reducing the input into the latent space, and the decoder reconstructs the latent codes of specific training samples to their original format.  The encoder can learn the fundamental features of the correlated inputs, and the decoder is trained to minimize the loss of information by reconstructing the original training samples.  These types of models can be viewed as a least squares regression problem formulated as 
\begin{equation}
L(\theta, \phi) = \frac{1}{N}\sum_{i=1}^{N}\|\textbf{x}_i - D (E (\textbf{x}_i|\phi)| \theta)\|_2^2,
\end{equation}
where $\textbf{x}_i$ is the $i^{th}$ training sample, $E(\cdot)$ is the output of the encoder function given a set of hyperparameters $\phi$, and $D(\cdot)$ is the output of the decoder function given a set of hyperparameters $\theta$.  Minimizing the autoencoder loss function $L(\theta, \phi)$) helps identify the optimal model parameters.  This learning process is unsupervised as the outputs are the same as the inputs, and functions by minimizing reconstruction error through back-propagation.  There are variants of the standard autoencoder, known as VAE \cite{kingma2019}, that involve enforcing a Gaussian distribution on the latent space, allowing for continuous sampling and interpolation of the latent space.  This permits a smooth transition between features, which is useful when new design instances are generated from the latent space.

Similar to autoencoders, GANs also use two neural networks for dimensionality reduction, a generator and a discriminator \cite{goodfellow2014}.  The generator is used to create synthetic samples of the training dataset, and the discriminator is used to differentiate between real and generated samples.  This creates an adversarial game-like scenario where both models are learning to try and outperform each other.  The training process follows a zero-sum game modeled as the value function  
\begin{equation}
V(G, D) = \mathbb{E}_{\textbf{z} \sim p_{\text{data}}}[\log D(\textbf{z})] + \mathbb{E}_{\textbf{x} \sim p_x}[\log(1 - D(G(\textbf{x})))],
\label{eqn:mutual_info}
\end{equation}
where $\mathbb{E}$ is the expectation operator, $\textbf{z}$ is a sample of real data distribution $p_{\text{data}}$ used to train the discriminator, $\textbf{x}$ is the latent vector sampled from distribution $p_x$, and $G(\textbf{x})$ is the generator network and $D(\textbf{z})$ is the discriminator network.  The advantage of GANs over VAEs is that they typically provide more high-quality samples (e.g., sharper images). However, these GANs also introduce challenges.  The training process is often complicated as it can run into problems like mode collapse, where the GAN only outputs a small number of samples.  There are also no controls for determining the specific features outputted by these GAN models, which can be a significant risk when used for commercial purposes. 

As an alternative to GANs, diffusion models represent a class of unsupervised learning that, instead of directly mapping a latent vector to data, learn to reverse a gradual noising process. During training, these models progressively corrupt the data by adding Gaussian noise at each time step $t=1,\ldots,T$ and train a neural network to denoise the corrupted samples. At inference, the trained model transforms stationary Gaussian noise into high-quality, representative samples. The diffusion process is described by the forward diffusion equation given as 
\begin{equation}
    q(\textbf{x}_t|\textbf{x}_{t-1}) = \mathcal{N}\left(\sqrt{1-\beta_t} \textbf{x}_{t-1}, \beta_t \mathcal{I}\right), \quad t=1,\ldots,T,
    \label{eqn:f_dif}
\end{equation}
and the reverse diffusion equation given as
\begin{equation}
    p_\theta(\textbf{x}_{t-1}|x_t) = \mathcal{N}\left(\textbf{x}_{t-1} \mid \mu_\theta(\textbf{x}_t, t), \Sigma_\theta(\textbf{x}_t, t)\right), \quad t=1,\ldots,T,
    \label{eqn:b_dif}
\end{equation}
where $\mathcal{N}(\cdot)$ denotes a normal distribution, $x_0$ represents the original data sample, $x_t$ is the noisy sample after $t$ steps, $\beta_t$ is the noise variance introduced at step $t$, and $\mathcal{I}$ is the identity matrix. The functions $\mu_\theta$ and $\Sigma_\theta$ are learned neural network parameters that model the reverse denoising process. A key advantage of diffusion models is their more stable training compared to GANs, while still producing diverse and high-fidelity samples. However, a notable drawback is their higher computational cost, requiring tens to hundreds of iterative denoising steps to generate each sample.


 To enable control of the generated features, diffusion models and GANs can be equipped with an auxiliary network. The addition of an auxiliary network enables control over generated features by encouraging the model to associate specific latent codes with semantically meaningful attributes in the output. This control is achieved by maximizing the mutual information between the latent codes and the generated outputs \cite{chen2016}, ensuring that variations in the latent space correspond to predictable changes in the generated features \cite{yu2019}. Moreover, using the latent space, where the controllable variables are learned by an encoder, as input to the GAN or diffusion model, enables the generation of high-fidelity samples with similar controllability as that provided by VAEs.  This helps prevent mode collapse and lack of feature control in GANs and diffusion models. In this work we use Info-GANs but the presented framework is compatible with both GANs and diffusion models.

\subsection{Multi-Objective Bayesian Optimization}
\label{sec:mobo}
The general formulation of design problems with multiple objectives can be expressed as
\begin{equation}
    \textbf{x}^*=\argmax_{x\in\chi} f_i(\textbf{x}),\quad i=1,\ldots,q,
    \label{obj}
\end{equation}
where $\textbf{x}$ is a $d$-dimensional vector of design variables that resides in a space of admissible designs $\chi\in\mathbb{R}^d$ (i.e., $\textbf{x}\in\chi\in\mathbb{R}^d$), the objective functions are given by $f_i|\chi \rightarrow \mathbb{R}$ (e.g., simulation models or physical experiments), and $q$ is the number of objective functions. In addition, engineers typically rely on the use of emulators when objective function evaluations are time or monetary-demanding activities. These emulators $\hat{f}_i|\chi \rightarrow \mathbb{R},\quad i=1,\ldots,q$ provide a fast way to evaluate an approximation of the true objective functions (i.e., $\hat{f}_i(\cdot)\approx f_i(\cdot),\quad i=1,\ldots,q$). The problem then becomes to optimize Eqn.~\ref{obj} and make inference into $\hat{f}_i,\quad i=1,\ldots,q$ concurrently.

A common choice of emulation models for design are Gaussian processes (GP) as they are data efficient and provide a posterior predictive distribution \cite{rasmussen2003}. The latter is important as it enables adaptive evaluation and efficient optimization of the objective functions \cite{vanbeek2021}. The primary assumption that underpins a GP is that any two observations $f_i(\textbf{x}),\quad i=1,\ldots,q$ and $f_i(\textbf{x}^{\prime}),\quad i=1,\ldots,q$ are jointly normally distributed. Under this assumption, we can model a GP of the $i^{th}\quad i=1,\ldots,q$ output as $\hat{f}_i(\textbf{x})$ that follows a multivariate Gaussian distribution $\mathcal{N}\left(m_i(\textbf{x}),  c_i(\textbf{x},\textbf{x}^{\prime})\right)$, with a prior mean function $m_i(\cdot)$ and a covariance function $c_i(\cdot,\cdot)$. The prior mean is often modeled through a basis expansion \cite{vanbeek2024}, but recent work has shown promising results with more flexible functions such as neural networks \cite{mora2024}. In addition, the covariance function is typically modeled to be square exponential $c_i(\textbf{x},\textbf{x}^{\prime})=\sigma_i^2exp\left( 
- \sum_{j=1}^d 10^{\omega_j^{(i)}}(x_j-x_j^{\prime})^2 \right)$ \cite{rasmussen2003},  where $\boldsymbol{\omega}^{(i)} = \left\{\omega_1^{(i)},\ldots,\omega_d^{(i)}\right\}^T$ and $\sigma_i^2$ are hyperparameters that need to be inferred from the training data (e.g., maximum likelihood approximation, full Bayesian, or k-fold cross validation). Finally, by conditioning the GP of the $i^{th}$ objective function on an observed set of $n$ training samples with inputs $\textbf{X}=\left\{\textbf{x}_,\ldots,\textbf{x}_n \right\}^T$ and outputs $\textbf{Y}_i=\left\{y^{(i)}_1,\ldots y^{(i)}_n \right\}^T$ we can get a posterior predictive distribution for any input $\textbf{x}_0\in \chi$ as
\begin{equation}
    \hat{f}_i(\textbf{x}_0)|\textbf{Y}_i\sim \mathcal{N}\left(\mu(\textbf{x}_0),s^2(\textbf{x}_0)\right),
\end{equation}
where $\mu(\textbf{x}_0)$ and $s^2(\textbf{x}_0)$ are the posterior predictive mean and variance, respectively \cite{forrester2009}.

To solve the optimization problem in Eqn.~\ref{obj} in a data-efficient manner, we can leverage the posterior predictive distribution of the GPs to identify what simulations/experiments should be conducted next. This is a process known as Bayesian optimization \cite{vanbeek2021} where we start with a relatively small set of space-filling samples (e.g., typically chosen to be $2\times d< n < 5\times d$ \cite{vanbeek2020a}). Subsequently, additional samples are identified by maximizing an acquisition function using the GP posterior predictive distribution conditioned on the available set of observations. Examples of acquisition functions for single objective problems include expected improvement \cite{forrester2009}, probability of improvement \cite{jones1998}, knowledge gradient \cite{frazier2016}, and entropy search \cite{hennig2012}. However, having a metric to measure the relative merit of conducting a specific simulation/experiment is more challenging when measured with respect to a $p$-dimensional set of Pareto optimal samples $\mathcal{P}\subset \left\{\textbf{y}_1^{1:q},\ldots,\textbf{y}_n^{1:q}\right\}^T \in \mathbb{R}^{p\times q}$ (i.e., multi-objective problems). One way that this can be achieved is by measuring the hypervolume improvement that a new observation $\textbf{y}_{0}^{(1:q)}$ would provide with respect to the current set of Pareto optimal samples given as
\begin{equation}
HVI\left(\mathcal{P},\textbf{y}_{0}^{(1:q)},\textbf{r}\right)=\lambda\left(\mathcal{P}\cup\textbf{y}_{0}^{(1:q)},\textbf{r}\right)-\lambda\left({\mathcal{P}},\textbf{r}\right)
\end{equation}
where $\lambda(\cdot)$ is the Lebesgue measure that is bounded from above by a reference point $\textbf{r}$ \cite{yang2019}. Using GP emulators, we know that $\textbf{y}_{0}^{(1:q)}$ is a random variable that follows a multivariate normal distribution given as $\textbf{y}_{0}^{(1:q)} = \left\{ \hat{f}_1(\textbf{x}_0)|\textbf{Y}_1,\ldots,\hat{f}_q(\textbf{x}_0)|\textbf{Y}_q \right\}^T$. Consequently, a new sampling location $\textbf{x}_{new}$ can be identified by maximizing the hypervolume and taking the expectation with respect to the uncertainty in $\textbf{y}_{0}^{(1:q)}$ as
\begin{align}
\textbf{x}^*_{new} & = \argmax_{\textbf{x}_{new}\in\chi} \mathbb{E}_{\sim \textbf{y}_{0}^{(1:q)}} HVI\left(\mathcal{P},\textbf{y}_{0}^{(1:q)},\textbf{r}|\textbf{X},\textbf{Y}\right),\label{eqn:aq} \\
& = \argmax_{\textbf{x}\in\chi}\alpha(\textbf{x}|\textbf{X},\textbf{Y}), \nonumber
\end{align}
where $\alpha(\cdot)$ is referred to as the acquisition function that can be optimized with respect to $\textbf{x}$ through $\textbf{y}_{0}^{(1:q)}$ using conventional numerical solvers. Finally, the new sampling location $\textbf{x}_{new}$ can be used as an input to the simulation model or physical experiments to update the training data set with a new sample $\left\{\textbf{x}_{new},f(\textbf{x}_{new}\right\}$. Iteratively updating the GPs with additional training samples will provide the designer with an approximated Pareto frontier of progressively higher accuracy.




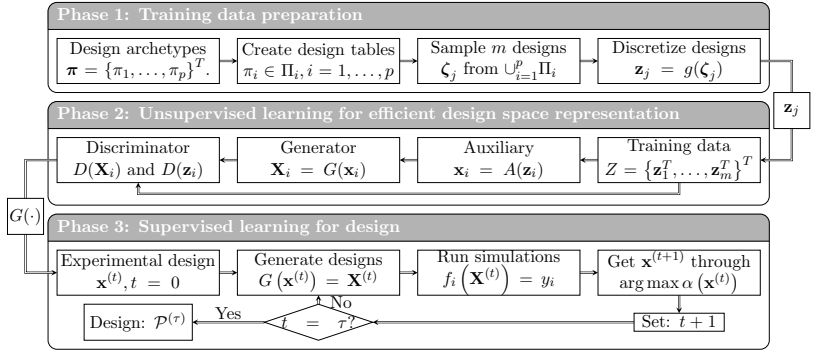
\begin{figure*}
    \centering
    \resizebox{\linewidth}{!}{
    \begin{tikzpicture}

    \filldraw [color=black!30!white, fill=black!30!white, rounded corners=2mm] (10mm,79mm) rectangle (170mm,85mm);
    \filldraw [color=black!30!white, fill=black!30!white] (10mm,79mm) rectangle (170mm,83mm);
    \draw [color=black, rounded corners=2mm](10mm,65mm) rectangle (170mm,85mm);
    \draw [color=black](10mm,79mm) -- (170mm,79mm);
    \node[anchor=center, align=left, text width = 156mm, color = white] at (90mm,82mm) {\textbf{Phase 1: Training data preparation}};
    
    \filldraw [color=black!30!white, fill=black!30!white, rounded corners=2mm] (10mm,57mm) rectangle (170mm,63mm);
    \filldraw [color=black!30!white, fill=black!30!white] (10mm,57mm) rectangle (170mm,61mm);
    \draw [color=black, rounded corners=2mm](10mm,40mm) rectangle (170mm,63mm);
    \draw [color=black](10mm,57mm) -- (170mm,57mm);
    \node[anchor=center, align=left, text width = 156mm, color = white] at (90mm,60mm) {\textbf{Phase 2: Unsupervised learning for efficient design space representation}};
    
    \filldraw [color=black!30!white, fill=black!30!white, rounded corners=2mm] (10mm,32mm) rectangle (170mm,38mm);
    \filldraw [color=black!30!white, fill=black!30!white] (10mm,32mm) rectangle (170mm,36mm);
    \draw [color=black, rounded corners=2mm](10mm,8mm) rectangle (170mm,38mm);
    \draw [color=black](10mm,32mm) -- (170mm,32mm);
    \node[anchor=center, align=left, text width = 156mm, color = white] at (90mm,35mm) {\textbf{Phase 3: Supervised learning for design}};

    \draw [->,double, -stealth](48mm,72mm) -- (52mm,72mm);  
    \draw [->,double, -stealth](88mm,72mm) -- (92mm,72mm);  
    \draw [->,double, -stealth](128mm,72mm) -- (132mm,72mm);

    \draw [->,double, -stealth](168mm,72mm) -- (175mm,72mm) -- (175mm,50mm) -- (168mm,50mm); 
    \filldraw [color=black, fill=white] (171mm,57mm) rectangle (179mm,65mm);
    \node[anchor=center, align=center, text width = 36mm] at (175mm,61mm) {$\textbf{z}_j$};
    
    \filldraw [color=black, fill=white] (12mm,67mm) rectangle (48mm,77mm);
    \node[anchor=center, align=center, text width = 36mm] at (30mm,72mm) {Design archetypes $\boldsymbol{\pi}=\left\{\pi_1,\ldots,\pi_p \right\}^T$.};

    \filldraw [color=black, fill=white] (52mm,67mm) rectangle (88mm,77mm);
    \node[anchor=center, align=center, text width = 36mm] at (70mm,72mm) {Create design tables \\ $\pi_i \in \Pi_i,i=1,\ldots,p$};

    \filldraw [color=black, fill=white] (92mm,67mm) rectangle (128mm,77mm);
    \node[anchor=center, align=center, text width = 36mm] at (110mm,72mm) {Sample $m$ designs $\boldsymbol{\zeta}_j$ from $\cup_{i=1}^p \Pi_i$ };

    \filldraw [color=black, fill=white] (132mm,67mm) rectangle (168mm,77mm);
    \node[anchor=center, align=center, text width = 36mm] at (150mm,72mm) {Discretize designs  \\ $\textbf{z}_j=g(\boldsymbol{\zeta}_j)$};

    \draw [<-,double, -stealth](52mm,50mm) -- (48mm,50mm);  
    \draw [<-,double, -stealth](92mm,50mm) -- (88mm,50mm);  
    \draw [<-,double, -stealth](132mm,50mm) -- (128mm,50mm);  
    \draw [<-,double, -stealth](150mm,45mm) -- (150mm,42.5mm) -- (30mm,42.5mm) -- (30mm,45mm);  
    \draw [<-,double, -stealth](12mm,50mm) -- (5mm,50mm) -- (5mm,25mm) -- (12mm,25mm);  
    \filldraw [color=black, fill=white] (1mm,33.5mm) rectangle (9mm,41.5mm);
    \node[anchor=center, align=center, text width = 36mm] at (5mm,37.5mm) {$G(\cdot)$};

    \filldraw [color=black, fill=white] (12mm,45mm) rectangle (48mm,55mm);
    \node[anchor=center, align=center, text width = 36mm] at (30mm,50mm) {Discriminator \\ $D(\textbf{X}_i)$ and $D(\textbf{z}_i)$};

    \filldraw [color=black, fill=white] (52mm,45mm) rectangle (88mm,55mm);
    \node[anchor=center, align=center, text width = 36mm] at (70mm,50mm) {Generator\\ $\textbf{X}_i=G(\textbf{x}_i)$};

    \filldraw [color=black, fill=white] (92mm,45mm) rectangle (128mm,55mm);
    \node[anchor=center, align=center, text width = 36mm] at (110mm,50mm) {Auxiliary \\ $\textbf{x}_i=A(\textbf{z}_i)$};

    \filldraw [color=black, fill=white] (132mm,45mm) rectangle (168mm,55mm);
    \node[anchor=center, align=center, text width = 36mm] at (150mm,50mm) {Training data \\ $Z=\left\{\textbf{z}_1^T,\ldots,\textbf{z}_{m}^T \right\}^T$};

    \draw [->,double, -stealth](48mm,25mm) -- (52mm,25mm);  
    \draw [->,double, -stealth](88mm,25mm) -- (92mm,25mm);  
    \draw [->,double, -stealth](128mm,25mm) -- (132mm,25mm);  
    \draw [->,double, -stealth](150mm,20mm) -- (150mm,16mm); 
    \draw [->,double, -stealth](140mm,14mm) -- (82mm,14mm); 
    \draw [->,double, -stealth](58mm,14mm) -- (42mm,14mm); 
    \draw [->,double, -stealth](70mm,14mm) -- (70mm,20mm);

    \filldraw [color=black, fill=white] (12mm,20mm) rectangle (48mm,30mm);
    \node[anchor=center, align=center, text width = 36mm] at (30mm,25mm) {Experimental design \\ $\textbf{x}^{(t)}, t=0$};

    \filldraw [color=black, fill=white] (52mm,20mm) rectangle (88mm,30mm);
    \node[anchor=center, align=center, text width = 36mm] at (70mm,25mm) {Generate designs \\ $G\left(\textbf{x}^{(t)}\right) = \textbf{X}^{(t)}$};

    \filldraw [color=black, fill=white] (92mm,20mm) rectangle (128mm,30mm);
    \node[anchor=center, align=center, text width = 36mm] at (110mm,25mm) {Run simulations \\ $f_i\left(\textbf{X}^{(t)}\right) = y_i$};

    \filldraw [color=black, fill=white] (132mm,20mm) rectangle (168mm,30mm);

    \node[anchor=center, align=center, text width = 45mm] at (150mm,25mm) {Get $\textbf{x}^{(t+1)}$ through\\ $\argmax\alpha\left(\textbf{x}^{(t)}\right)$};

    \filldraw [color=black, fill=white] (70mm,10mm) -- (58mm,14mm) -- (70mm,18mm) -- (82mm,14mm) -- cycle;
    \node[anchor=center, align=center, text width = 60mm] at (70mm,14mm) {$t = \tau$?};

    \filldraw [color=black, fill=white] (140mm,12mm) rectangle (160mm,16mm);
    \node[anchor=center, align=center, text width = 20mm] at (150mm,14mm) {Set:  $t+1$};

    \filldraw [color=black, fill=white] (18mm,10mm) rectangle (42mm,18mm);
    \node[anchor=center, align=center, text width = 26mm] at (30mm,14mm) {Design: $\mathcal{P}^{(\tau)}$};
    
    \node[anchor=center, align=center, text width = 20mm] at (75mm,18.5mm) {No};
    \node[anchor=center, align=center, text width = 20mm] at (50mm,16mm) {Yes};

    \end{tikzpicture}
    }
    \caption{A three-phased approach for Systematic Generation of Training Data and optimization for problems with heterogeneous design variables.}
    \label{fig:flowchart}
\end{figure*}

\section{Framework}
To address the complexity of multi-objective design optimization with heterogeneous inputs, we propose a three-phase framework. 

\subsection{A Systematic Framework for Design with Heterogeneous Data}
This section introduces a framework to help designers navigate heterogeneous design spaces through systematic training sample generation, unsupervised learning with an Info-GAN, and design optimization via Bayesian optimization, as illustrated in Figure~\ref{fig:flowchart}. In Phase 1, new design instances (e.g., technical drawings) are generated from a heterogeneous set of variables. Subsequently, in Phase 2 we employ these design instances to train an Info-GAN, reducing and unifying the dimensionality of competing design inputs. Finally, in Phase 3, the resulting latent space is used to establish an iterative loop, leveraging physical or simulation experiments to evaluate and optimize new designs. Despite its potential, this framework presents practical challenges, particularly in generating a robust and representative training dataset required for unsupervised learning. 

This framework applies to many design problems; for this paper, we explore the optimization of flow battery manifolds as an illustrative problem.  This example will provide a practical application and showcase the benefits of this framework, both in terms of effectiveness and efficiency. Flow batteries pose a complex problem, with a near-infinite number of possible manifold configurations and multiple optimization functions, including maximizing charge capacity and minimizing the average charge voltage of the batteries.

\subsection{Phase 1: Training Set Generation}
Creating a training dataset for an Info-GAN model is the first phase of the introduced design framework.  Training sets exist for many systems, such as the University of Illinois Urbana-Champagne airfoil database\cite{uiuc2024}. However, for many engineering challenges such libraries are not available. Consequently, we need to establish a training set with appropriate size and complexity to prevent the discriminator from over-fitting for which a few thousand images will be required \cite{karras2020}. To this end, we suggest identifying a set of design archetypes $\boldsymbol{\pi}=\left\{\pi_1,\ldots,\pi_p \right\}^T$ that individually have homogeneous inputs but jointly have heterogeneous inputs. Subsequently, the admissible design space can be identified for each archetype using conventional methods of experimental design. 

For flow batteries, previous work has been done exploring different manifold designs, providing insight into a reasonable design space \cite{oconnor2022}.  Based on this, we decided on five designs: two-prong, three-prong, four-prong, five-prong, and a diffuser with three inner geometries, shown in Table \ref{tab:designs}.  These five different designs provide sufficient variation in the design space to allow for a deep understanding of the effects of various design features but will keep the training process simple enough to be trained with limited available resources \cite{chen2020}, as we will show in Sec.~\ref{subsec:results_sim}. This present a challenge to select design archetypes that provide a uniform coverage of the space of admissible designs, a problem typically referred to as diversity-based actively learning \cite{Xie2023,lee2023}. While flow batteries are used as an illustrative example, this process could be applied to a broad range of design challenges, allowing researchers and industry to efficiently explore complex heterogeneous design spaces. 

\begin{table}[t]
\caption[Table]{Table of five design manifold archetypes}\label{tab:designs}
\centering
\begin{tabular}{llr}
\toprule
\textbf{Image} & \textbf{Description} \\
\midrule
\includegraphics[width=3cm]{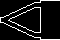} & 2 Prong Manifold \\
\includegraphics[width=3cm]{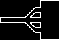} & 3 Prong Manifold \\
\includegraphics[width=3cm]{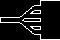} & 4 Prong Manifold \\
\includegraphics[width=3cm]{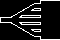} & 5 Prong Manifold \\
\includegraphics[width=3cm]{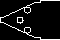} & Inner Geometry Manifold \\
\bottomrule
\end{tabular}
\end{table}

With the design archetypes identified, the next step is to create a design table for these archetypes.  For the flow battery case, we first create a base model of the different manifold archetypes, something that can be achieved in any computer-aided design (CAD) software.  For the available archetypes, we can use CAD tools to define the different dimensions that can be varied to generate new designs (e.g., for a specific design we can change the inlet angle and channel width). We generated 25 unique designs for each archetype, bringing the total design count to 125 unique manifolds.  Specifically, we used SOLIDWORKS (Dassault Systèmes, France)  with the ``Design Table'' feature, to generate different designs $\Pi_i,\quad i=1,\ldots,p$. These design tables can be established using conventional experimental design methods (e.g., Latin hypercube sampling, or factorial designs) to establish the necessary training data set $\mathcal{Z} = \left\{\boldsymbol{\zeta}_1^T,\ldots, \boldsymbol{\zeta}_{m}^T \right\}^T$.

\begin{figure}
    \centering
    \includegraphics[width=0.9\linewidth]{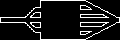}
    \caption{Mixed Design of 3-prong input and 5-prong output}
    \label{fig:mixed_design}
\end{figure}

From these design tables new designs can be generated in the form of images. However, such images need to have a consistent resolution, thus warranting the need for additional post-processing through a function $g(\cdot)$. For example, to represent the admissible design space, we exported these designs  $\mathcal{Z}$ from SOLIDWORKS as DXF files and mirrored them over the vertical axis (i.e., doubling the size of the generated training data set).  With these copies, we mix each initial left half with every mirrored copy.  An example of this is shown in Figure~\ref{fig:mixed_design}.  This provided a training dataset with a wide variety of possible inlet and outlet manifold configurations.  To convert this to something appropriate to train the Info-GAN, we need to convert these DXFs into a PNGs for training. We settled on $96 \times 96$ pixel images for training as they provided enough fidelity while reducing the computational cost. If the generated images are binary, then we recommend recoloring to RGB to add more information improving the training process.  The red regions indicated areas where no electrolytic fluid would flow.  The green regions indicated areas where there would be fluid flow through the manifold, and the blue regions indicated boundary layers that would surround the green areas.  This gave the Info-GAN more information than just the boundary walls to encode the electrode cavity design information in the color layers.  

\subsection{Phase 2: Unsupervised Learning}
The second phase of the framework involves using the training data set generated in Phase 1 (i.e., $Z=\left\{\textbf{z}_1^T,\ldots,\textbf{z}_{m}^T \right\}^T$) to train an info-GAN.  This is a joint process where the generator network $G(\cdot)$, the discriminator network $D(\cdot)$ and the auxiliary network $A(\cdot)$ are trained simultaneously to reduce the risk of mode collapse and make interpolation between designs possible \cite{lazarou2020}. When setting up these models, a designer will have to define the number of training epochs and the dimension of the latent space (i.e., they need to select a value for $d$). Concerning the latter, the variable $d$ determines how far the auxiliary network will reduce the input images and the available data that the GAN can use to reconstruct the original training samples. For our test case, we used an auxiliary network that reduces the initial samples down to 256, then 128, and then to our desired latent space dimension $d$.  This was done with fully connected  layers, using the LeakyReLU activation function and batch normalization. The purpose of the auxiliary network is to provide structure into the space of admissible design by maximizing the mutual information in the Info-GAN. 

To test the validity of the selected training parameters, designers typically have to empirically try different settings. For the flow battery example, we used a power of two to set the size of the latent space, where we trained the model for $d=4$, $8$, and $16$.  This is important as a smaller latent space will ensure that the models can encode a sufficient amount of information increasing its size will improve the reconstruction accuracy.  We determined empirically that $10,000$ epochs brought us to a learning plateau, and this was confirmed by comparing the generated outputs to standard inputs every 10 training epochs. To test the accuracy of the reconstructed images, we used a pixel comparison of their root mean squared error (RMSE), between the training images with the reconstruction from that same image.  The results for the different latent space dimensions have been shown in Table \ref{tab:RMS_Comparison}.  Based on this, we determined the optimal latent size is eight latent variables, keeping the simulation values low while maintaining similar RMSE values to the 16 latent variables, but had significant improvement over the four-dimensional latent space. 

\begin{table}[t]
\caption[Table]{Comparison of Latent Space Dimension and RMSE Values}\label{tab:RMS_Comparison}
\centering
\begin{tabular}{ll}
\toprule
\textbf{Latent Dim} & \textbf{Average RMSE} \\
\midrule
4 Latent Variables & 12.34\\
8 Latent Variables & 10.89\\
16 Latent Variables & 11.27\\
\bottomrule
\end{tabular}
\end{table}

\begin{figure}
    \centering
    \includegraphics[width=1\linewidth]{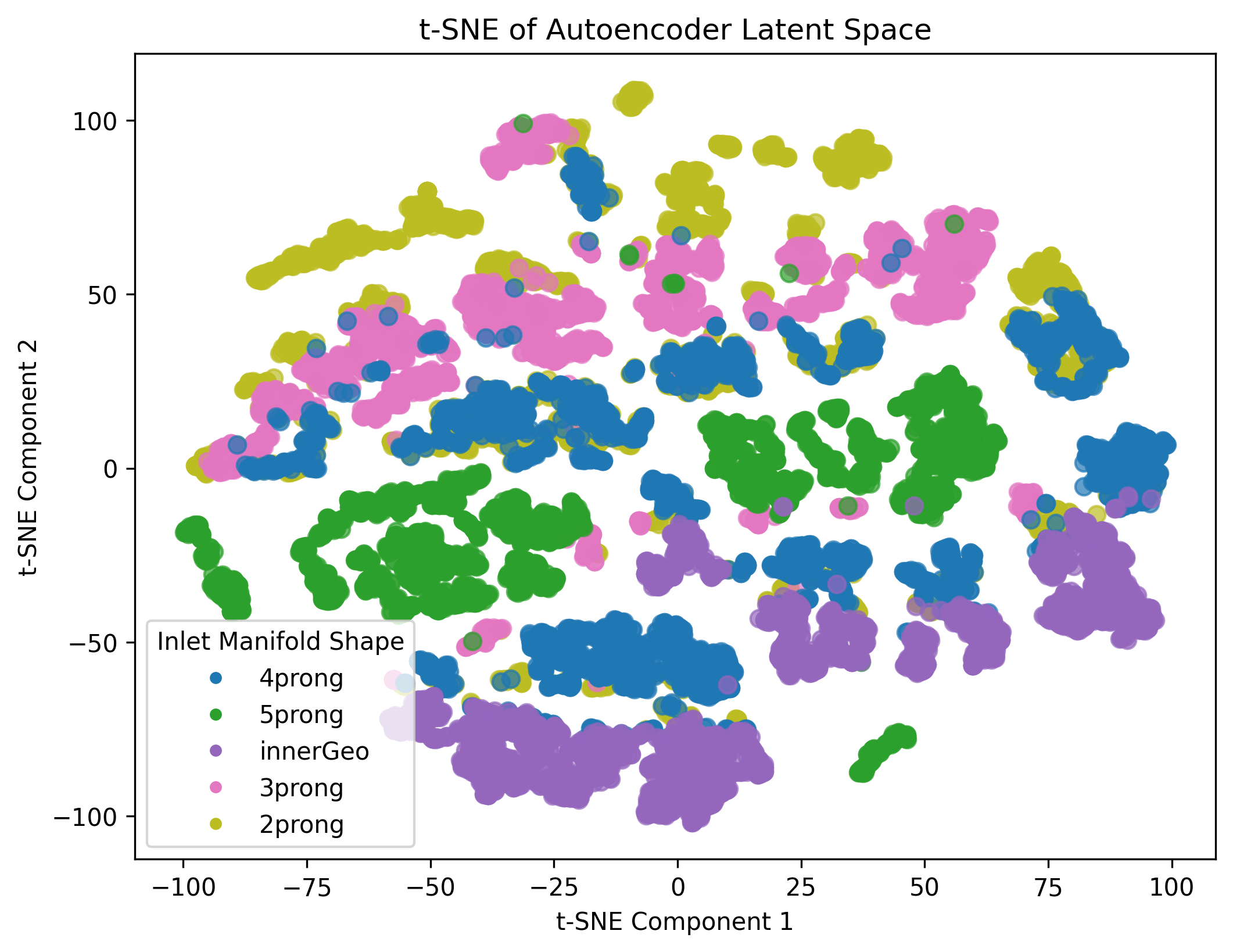}
    \caption{T-SNE Plot of Latent Space}
    \label{fig:tsne}
\end{figure}

For the Info-GAN with an 8-dimensional latent space, we wished to validate the training accuracy by visualizing the latent space.  To this end, we use a t-distributed stochastic neighbor embedding (T-SNE) plot as a way to visualize the learning process.  The role of the T-SNE plot is to reduce the dimensionality of the latent space into a two- or three-dimensional plot.  In these plots, it is expected to see distinct clusters of designs with different features, indicating that there are distinct groups with common characteristics.  In Figure~\ref{fig:tsne}, all of the test images are passed through the auxiliary network and then through the T-SNE function.  In addition, we colored the reconstructions based on their inlet manifolds based on their associated design archetype.  From this plot, we can observe that there are distinct clusters of data, where manifolds from the same archetype are clustered together. This is an encouraging observation as it shows that the trained model can distinguish between the different design features. 

\subsection{Phase 3: Design of Experiments}
\label{DoE}
To efficiently explore the design space, we use the multi-objective Bayesian optimization method introduced in Sec.~\ref{sec:mobo}.  To initiate this process, we need to have a Design of Experiments (DoE) that defines the first set of designs to be simulated $\textbf{x}^{(0)}$, where we used the superscript $(0)$ to indicate that this was the initial batch of experiments.  Because the training process of the Info-GAN forces the latent space to be a multivariate standard normal distribution, we need DoE to be in a hyper-sphere with a specific radius. For this purpose, we recommend the use of a Sobol sequence \cite{Sobol1967} from which a large array of space-filling samples can be generated. Subsequently, samples that fall outside of the hyper-circle can be removed until the desired number of initial training samples has been found.  This helps the GAN model because it is set to learn the latent space as a multivariate standard normal distribution.  One of the main benefits of the Sobol sequence generation is that it provides a more uniform distribution across multidimensional spaces than other options like pseudo-random generation processes.  That is to say, this approach provides a way to uniform coverage of the admissible designs in the latent space. 

The generated latent codes can then be passed through the generator network to create the image of the potential flow channels as $G\left(\textbf{x}^{(t)}\right) = \textbf{X}^{(t)}$, Subsequently, these flow channels are to be tested $f_i\left(\textbf{X}^{(t)}\right) = y_i,\quad,i=1,\ldots,q$, where $f_i(\cdot)$ using $q$ physical or simulation experiments. Next, we can use the data set $\textbf{D}^{(t)}=\left\{\textbf{X}^{(t)},\textbf{Y}^{(t)}_1,\ldots,\textbf{Y}^{(t)}_q\right\}$ to train the GP emulators and identify what experiments to conduct next $\textbf{x}^{(t+1)}$ by maximizing the expected hypervolume improvement acquisition function $\textbf{x}^{(t+1)}=\argmax\alpha\left(\textbf{x}^{(t)}|\textbf{D}^{(t)}\right)$ as defined in Eqn.~\ref{eqn:aq}. Note that batches containing multiple experiments can be allocated using a pre-posterior analysis \cite{vanbeek2021}. Finally, we proceed by updating the iteration counter by setting $t+1$ and by verifying if we have exhausted our experimental budget $\tau$ (i.e., check if $t<\tau$). If the training budget has not yet been exhausted, then the new batch of training samples $\textbf{x}^{(t+1)}$ will be fed through the generator and we continue the optimization loop. Alternative to using the experimental budget as a stopping criterion, a designer can also use the expected hypervolume improvement as an indicator to decide when to stop conducting more experiments. 

Through the above description, we solve an optimization problem that is different from the one presented in Sec.~\ref{sec:mobo}. Specifically, we solved the following problem
\begin{align}
    find: \quad & \textbf{x},& \label{obj2}\\
    min: \quad  & f_i(\textbf{x}), &i=1,\ldots,q, \nonumber \\
    s.t.: \quad  &||\textbf{x}||_{\mathcal{L}_2}\leq 2
    ,\nonumber
\end{align}
where $g_i(\cdot)$, $h_j(\cdot)$ are inequality and equality constraints. Finally, the design bounds $||\textbf{x}||_{\mathcal{L}_2}\leq 2$ ensure that new designs are generated within the latent space for which appropriate designs can be generated.

\section{Results}
In this section we will provide an introduction into the simulation setup used to test different manifold designs and explore the obtained results.

\subsection{Simulation of flow Batteries}
Flow batteries are an increasingly important electrochemical energy storage technology, particularly suited for large-scale, long-duration storage of energy generated by intermittent renewable sources. The key working principle of a flow battery is that two liquid electrolytes circulate from individual tanks, each through one half-cell of an electrochemical cell, separated by a membrane that facilitates ion transfer to provide charge compensation during charging and discharging. Despite scalability advantages over Li-ion batteries, with predictions of lower costs for durations beyond eight hours with aqueous electrolytes \cite{bai2023}, flow batteries suffer from low energy and power densities due to the intrinsic nature of liquid electrolytes. Although investigations into alternative electrolytes \cite{zhang2025}, electrodes \cite{ahn2025}, and membranes \cite{pileri2025} are widespread, research into the optimization of cell design is less intense. 

Establishing a uniform distribution of electrolyte throughout the electrode is nonetheless known to be important for minimizing net losses in the system, and internal manifold design has been shown to affect this \cite{oconnor2022}. One barrier to conducting this type of research, however, is the monetary and time expenses associated with flow battery experimental testing in a wide variety of cell topologies. Here, computational simulation can provide a more efficient means of surveying the design space than the costly fabrication of a high number of cells. Prior works have sought to identify higher performing cavity shapes \cite{oconnor2022}, and compression ratios \cite{charvat2020}, although a holistic approach to iteratively improve the design of manifolds for ``flow-through'' cell designs are lacking compared to work on flow field designs for ``flow-over'' cells \cite{hao2022,guo2025}.

Here, we employ a transient finite element simulation using COMSOL Multiphysics (COMSOL, v6.3, Sweden), similar to that presented by Shah et al.\cite{shah2008}, that couples computational fluid dynamics with the Butler-Volmer equation (electrochemical reactions), but in three dimensions. Our model combines electron transfer kinetics derived from the formulation of the Butler-Volmer equation used by Knehr et al.\cite{knehr2012}, simple proton transfer across a polymer electrolyte membrane using dynamics described by Schloegl, and electrolyte flow via Brinkmann computational fluid dynamics equations. An early iteration of this model will soon be available in the doctoral thesis of Dr Hugh O’Connor \cite{oconnor2024}. Its main assumptions are that the fluids are incompressible, flow is laminar, the physical properties of the electrode, electrolyte and membrane are homogeneous, and that there are no side-reactions or active-species cross-over across the membrane. Charge curves generated by the model were subsequently used to calculate the average charge voltage as a proxy metric for voltage efficiency (given the computational expense involved with modeling both charge and discharge curves in every case). The total charge time for the system run at a constant current density of $50\ mA\ cm^{-2}$, flow rate of $10\ mL\ min^{-1}$, and electrolyte volume in each tank of $25 mL$, was used to determine the charge capacity as an indication of the electrolyte utilization of the system. In each case, the three-dimensional electrode cavity output from SOLIDWORKS was used as an input geometry and subsequently meshed within COMSOL with a user-controlled mesh of approximately $150,000$ elements.

\begin{figure}
    \centering
    \includegraphics[width=0.9\linewidth]{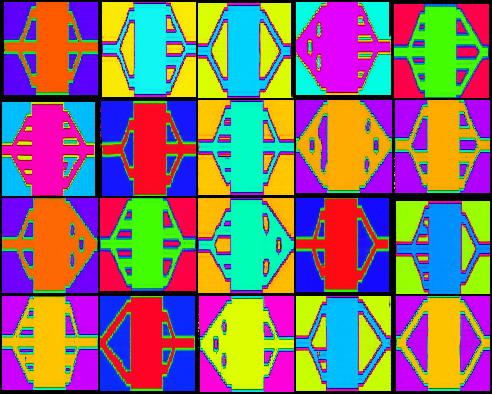}
    \caption{Initial experimental design containing 20 manifolds}
    \label{fig:doe}
\end{figure}

\begin{figure*}
    \centering
    \resizebox{\linewidth}{!}{
    \begin{tikzpicture}

    \node at (35mm,35mm) {\includegraphics[width=70mm]{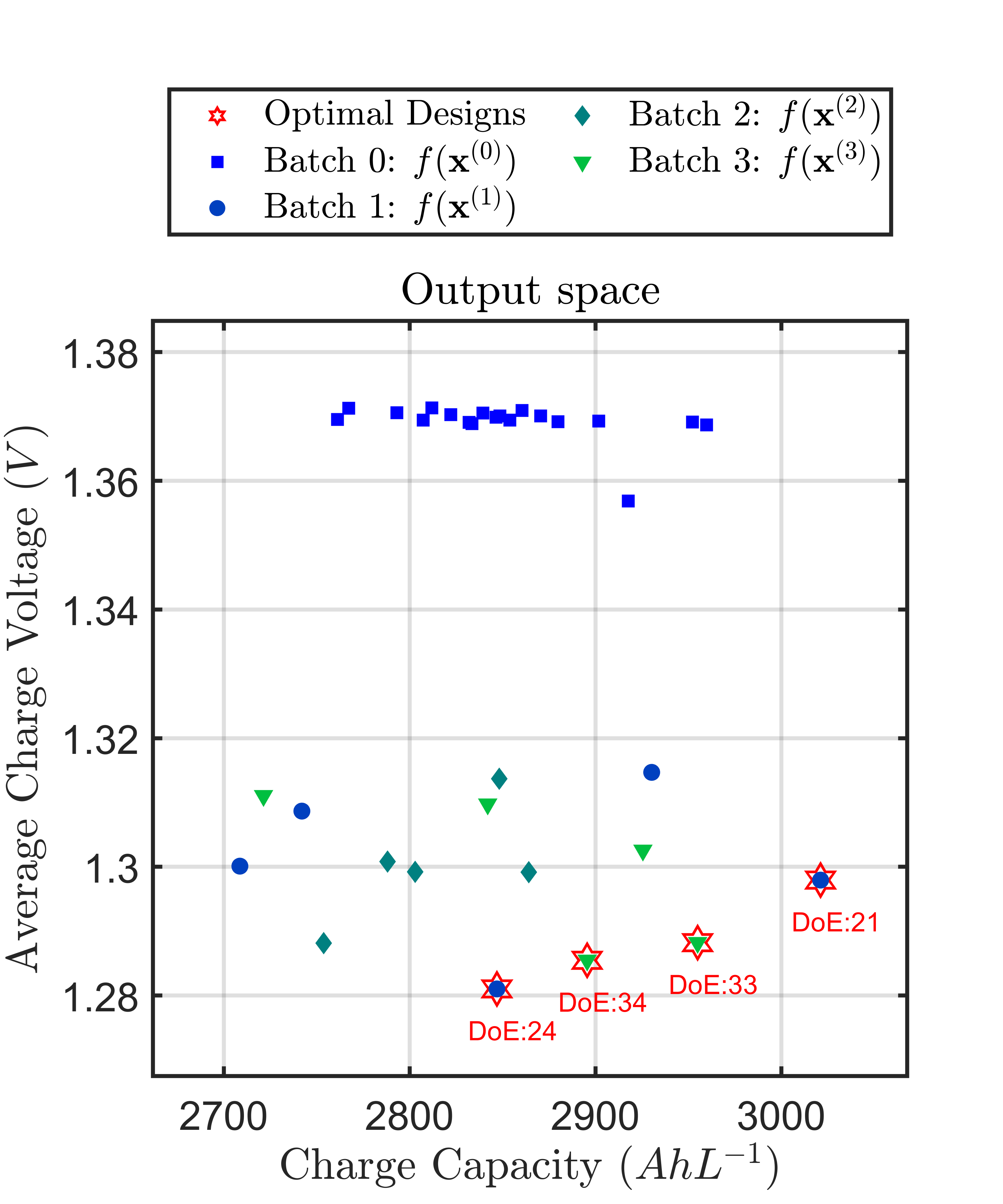}};

    \node at (85mm,52mm) {\includegraphics[width=30mm]{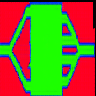}};
    \node at (120mm,52mm) {\includegraphics[width=30mm]{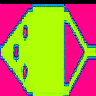}};
    \node at (85mm,15mm) {\includegraphics[width=30mm]{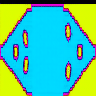}};
    \node at (120mm,15mm) {\includegraphics[width=30mm]{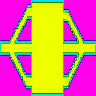}};

    \filldraw [color=black, fill=white] (3mm,68mm) rectangle (8mm,73mm);
    \node[anchor=center, align=center, text width = 20mm] at (5.5mm,70.5mm) {$A$};

    \filldraw [color=black, fill=white] (70mm,68mm) rectangle (87mm,73mm);
    \node[anchor=center, align=left, text width = 25mm] at (83mm,70.5mm) {B - DoE24};

    \filldraw [color=black, fill=white] (105mm,68mm) rectangle (122mm,73mm);
    \node[anchor=center, align=left, text width = 25mm] at (118mm,70.5mm) {C - DoE34};

    \filldraw [color=black, fill=white] (70mm,31mm) rectangle (87mm,36mm);
    \node[anchor=center, align=left, text width = 25mm] at (83mm,33.5mm) {D - DoE33};

    \filldraw [color=black, fill=white] (105mm,31mm) rectangle (122mm,36mm);
    \node[anchor=center, align=left, text width = 25mm] at (118mm,33.5mm) {E - DoE21};

    \end{tikzpicture}
    }
    \caption{Visualization of the approximated Pareto frontier evolution and the associated manifolds. A) Pareto frontier, and B) optimal design associated with DoE 24, C) optimal design associated with DoE 34, D) optimal design associated with DoE 33, and E) optimal design associated with DoE 21.}
    \label{fig:optimization_results}
\end{figure*}

\subsection{Optimization History and Findings}
\label{subsec:results_sim}
To initiate the optimization process for flow battery manifolds, we first generated an initial experimental design including 20 samples. The designs selected through the procedure described in Subsection~\ref{DoE} are shown in Figure\ref{fig:doe}, with the aspect ratio of the channels adjusted to form squares to improve visual comparison. From these initial designs, several notable observations can be made. First, the generated set of manifolds provides a well-distributed coverage of the space of admissible designs, incorporating a diverse range of manifold features (e.g., varying numbers of prongs and different types of inclusions). Additionally, we observe the emergence of novel features that were not present in the original training dataset $Z$. For example, we observed horizontal asymmetries (e.g., the manifold in the fifth column of the second row, the first column of the fifth row, and the fourth column of the fifth row), as well as new inclusion patterns (e.g., the wavy inclusion in the fourth column of the second row), and the presence of potentially incomplete prongs (e.g., the manifold in the first column of the second row). While these designs might not necessarily result in optimal performance, their diversity and physical realism indicate the validity of the generative model. The observation that our approach produces novel yet feasible designs suggests that the generative model effectively represents the space of admissible manifolds with structural plausibility.

Starting from the initial batch of experimental designs $\textbf{x}^{(0)}$, for which the simulation outputs are shown by the blue squares in Panel A of Figure~\ref{fig:optimization_results}. We build on the experimental design by using the aforementioned multi-objective Bayesian optimization method as outlined in Subsection~\ref{sec:mobo} to iteratively allocate a total of $\tau=3$ batches of five new experiments each $\textbf{x}^{(t)},\quad t=1,\ldots,\tau$. The history of the predicted performance of the new batches has been shown through the blue squares, azure circles, teal diamonds, green downward-facing triangles, and emerald upward-facing triangles in Figure~\ref{fig:optimization_results}. From these results, we can observe that the initial set of experiments had an appreciable variability in terms of charge capacity and the average charge voltage had only a minor variability relative to the total conducted set of simulation experiments. However, as additional observations are conducted, the Pareto frontier gets gradually pushed to the bottom right of the graph until the final set of optimized manifolds has been achieved as indicated by the experiments outlines by red stars in Figure~\ref{fig:optimization_results}. Finally, we have shown the Pareto optimal manifolds in Panels B to E of Figure~\ref{fig:optimization_results}.

It can be observed that the electrode cavity designs included in the initial batch that were asymmetric along the vertical axis are not represented in the designs associated with the Pareto frontier, in keeping with the symmetry of the multiphysics simulated. Nevertheless, two of the four designs (DoE24 and DoE34) on the frontier display clear asymmetry over the horizontal axis, whereas DoE21 and DoE 33 are almost symmetrical along this axis. DoE24 and DoE34 present lower average charge voltage, whereas DoE21 and DoE33 present greater charge capacity, somewhat at the expense of average charge voltage. It may be interpreted that to minimize average charge voltage, having output manifolds with exit channels that differ from the entry channels of the input manifold leads to less channeling of electrolyte through the electrode cavity and therefore may lead to a more uniform electrolyte velocity field and concomitantly lower mass transport limitations. Further work would be needed to verify this and validate the finding experimentally. Nevertheless, there is also a design steer from the symmetrical DoE21 and DoE33 topologies that entry of electrolyte at the extremities, as well as along the central axis (three prongs in total), whether by prismatic channels or through a manifold dotted with channel obstacles, may give rise to cells with greater charge capacity that designs with four or five prongs. Therefore, there appears to be a balance between the degree to which flow is impeded upon entry and the overall cavity symmetry when attempting to optimize for higher charge capacity and lower average charge voltage. 

\subsection{Interpretation of the Latent Space}
A concern with the use of Info-GANs for design is the lack of interpretability that originates from the use of latent space. Specifically, in conventional Info-GAN architectures, the latent variables provide a means to control the designs but have no interpretable meaning. However, such insight could be gleaned by connecting the latent space of variables to design features and performance. In the analysis of the Pareto frontier in Section~\ref{subsec:results_sim} we already studied the relation between the design features and their performance, but using the GP emulators we can extend this analysis to also include the latent variables. Specifically, consider that we sample $N$ realizations from the GPs' posterior predictive distributions for a set of $q$ candidate designs drawn from the latent space $\textbf{x}_{can} = \left\{ \textbf{x}_{1,can}^T,\ldots,\textbf{x}_{q,can}^T \right\}^T$ (e.g., through the use of a Sobol sequence). Then we can approximate the probability that a candidate design is Pareto optimal as
\begin{equation}
    P_{\textbf{x}^*}(\textbf{x}_{i,can}) \approx \frac{1}{N}\sum_{j=1}^N I\left(\textbf{y}_j(\textbf{x}_{i,can}) \in \mathcal{P}(\textbf{y}_j(\textbf{x}_{can}))\right),
    \label{eqn:mc_post}
\end{equation}
where $I(\cdot)$ is an identity function that equals to one when the $i^{th}$ candidate sample is Pareto optimal in the $j^{th}$ output $\textbf{y}_j(\cdot)$ sampled from the posterior predictive distribution. Note that Eqn.~\ref{eqn:mc_post} provides only an approximation of the probabilities as predicted by our GP emulators.

\begin{figure*}
    \centering
    \begin{small}
    \resizebox{\linewidth}{!}{
    \begin{tikzpicture}

    \node at (20mm,-40mm) {\includegraphics[width=50mm]{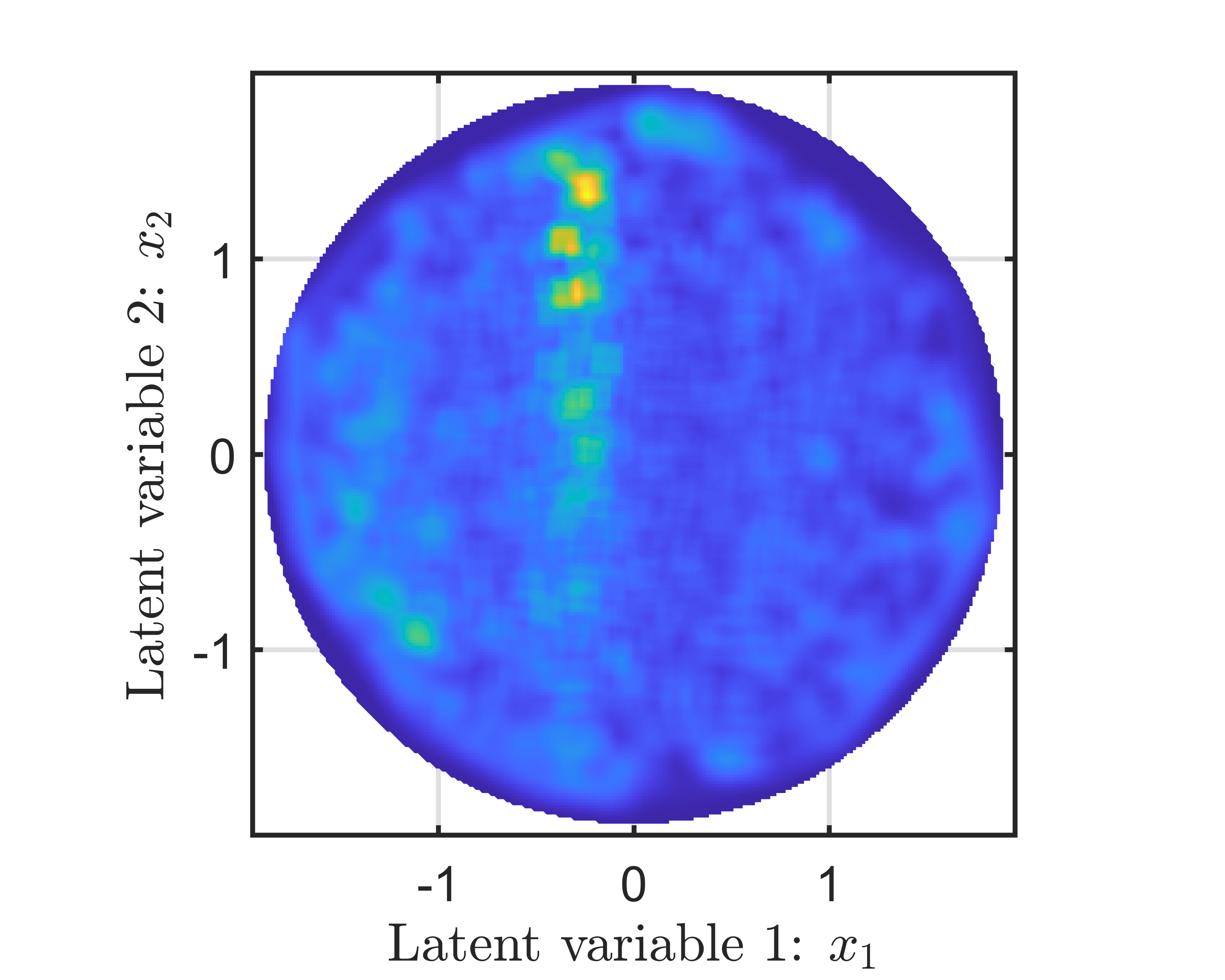}};
    \node at (65mm,0mm) {\includegraphics[width=50mm]{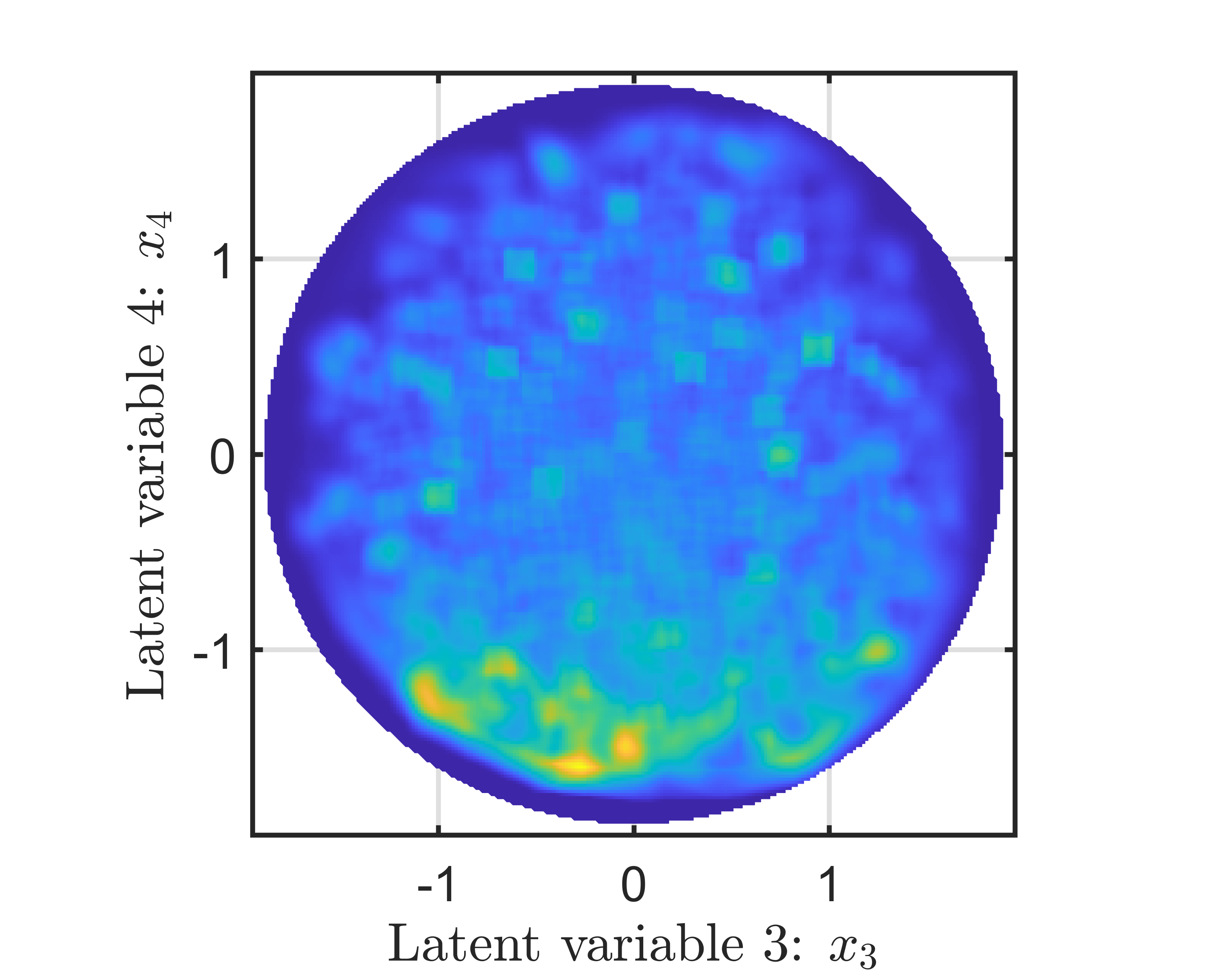}};
    \node at (20mm,0mm) {\includegraphics[width=50mm]{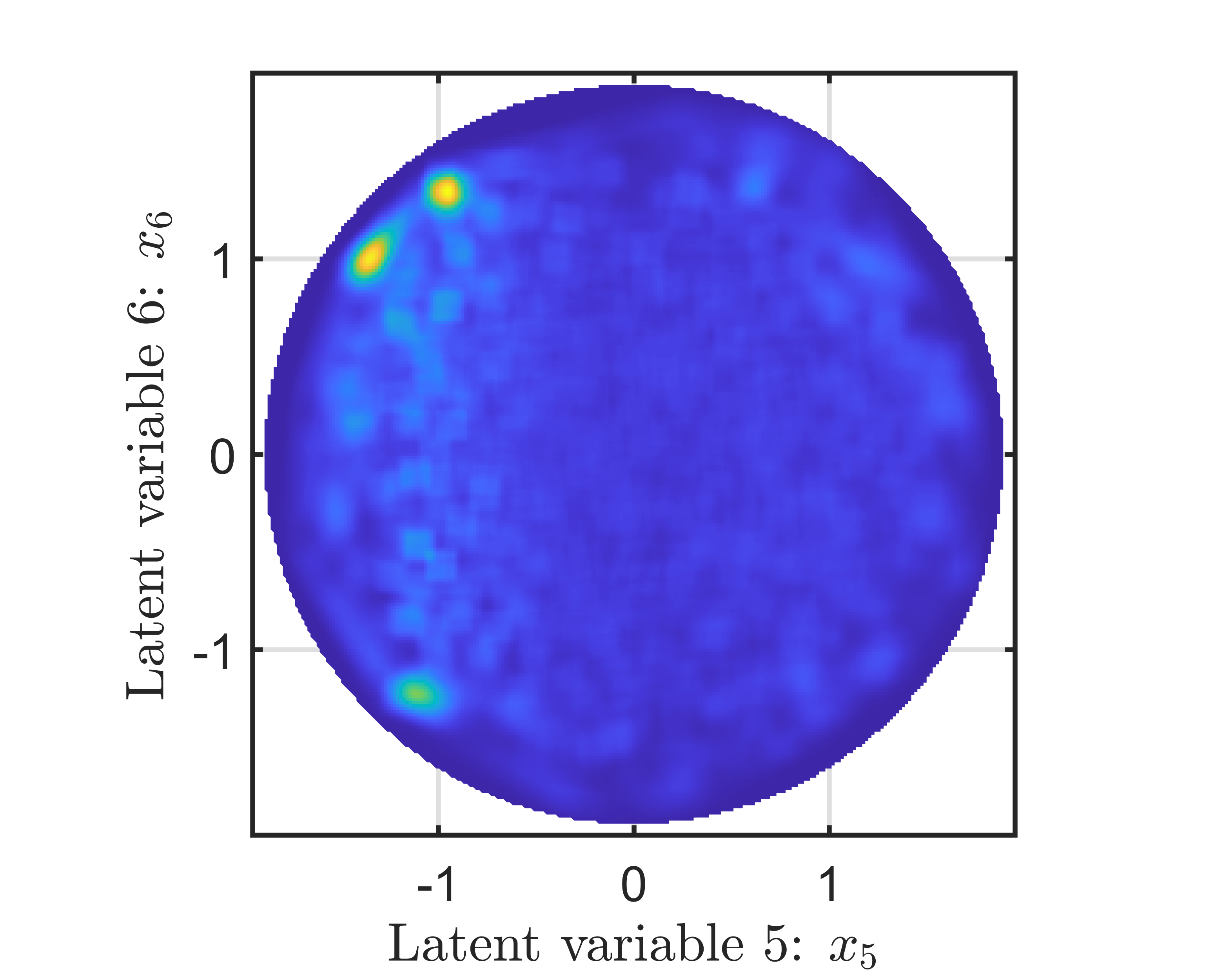}};
    \node at (65mm,-40mm) {\includegraphics[width=50mm]{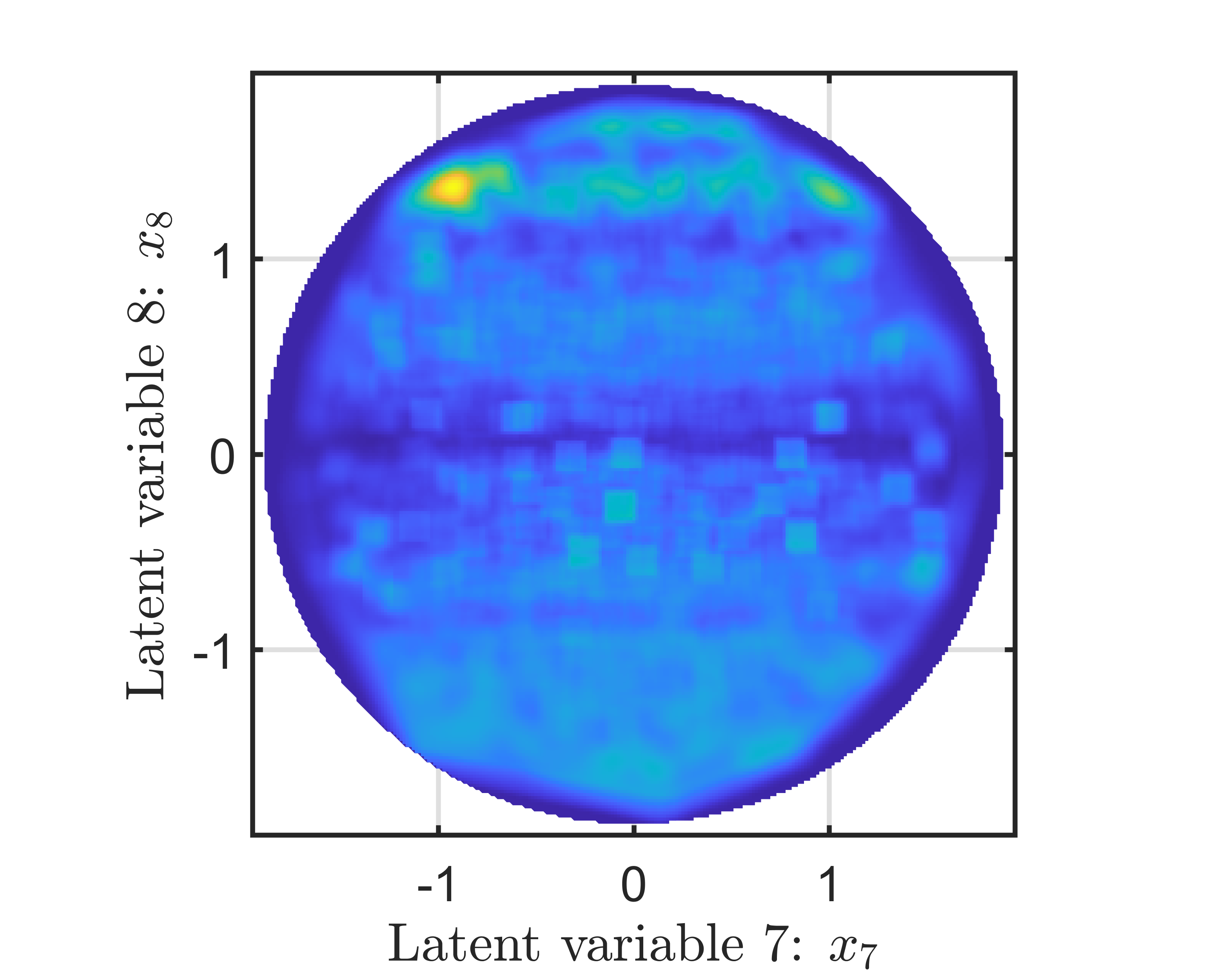}};
    
    \node at (42.5mm,25mm) {\includegraphics[width=80mm]{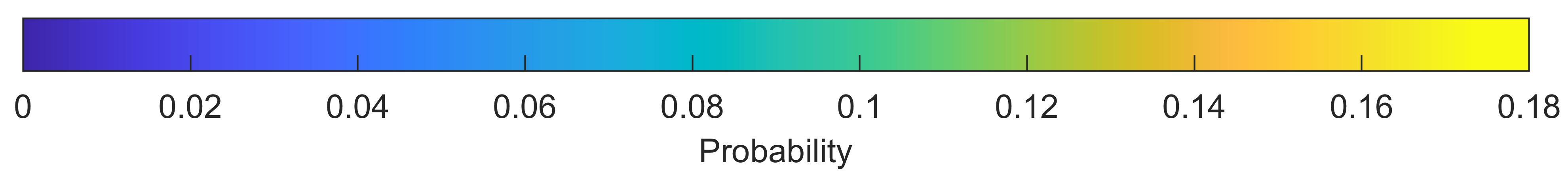}};

    \filldraw [color=black, fill=white] (34mm,13mm) rectangle (39mm,18mm);
    \node[anchor=center, align=left, text width = 5mm] at (37.5mm,15.5mm) {A};
    
    \filldraw [color=black, fill=white] (34mm+45mm,13mm) rectangle (39mm+45mm,18mm);
    \node[anchor=center, align=left, text width = 5mm] at (37.5mm+45mm,15.5mm) {B};

    \filldraw [color=black, fill=white] (34mm,13mm-40mm) rectangle (39mm,18mm-40mm);
    \node[anchor=center, align=left, text width = 5mm] at (37.5mm,15.5mm-40mm) {C};

    \filldraw [color=black, fill=white] (34mm+45mm,13mm-40mm) rectangle (39mm+45mm,18mm-40mm);
    \node[anchor=center, align=left, text width = 5mm] at (37.5mm+45mm,15.5mm-40mm) {D};
    

    
    \end{tikzpicture}
    }
    \end{small}
    \caption{Marginal distributions representing the probability that designs in the latent space are predicted to be Pareto optimal}
    \label{fig:latent_results}
\end{figure*}
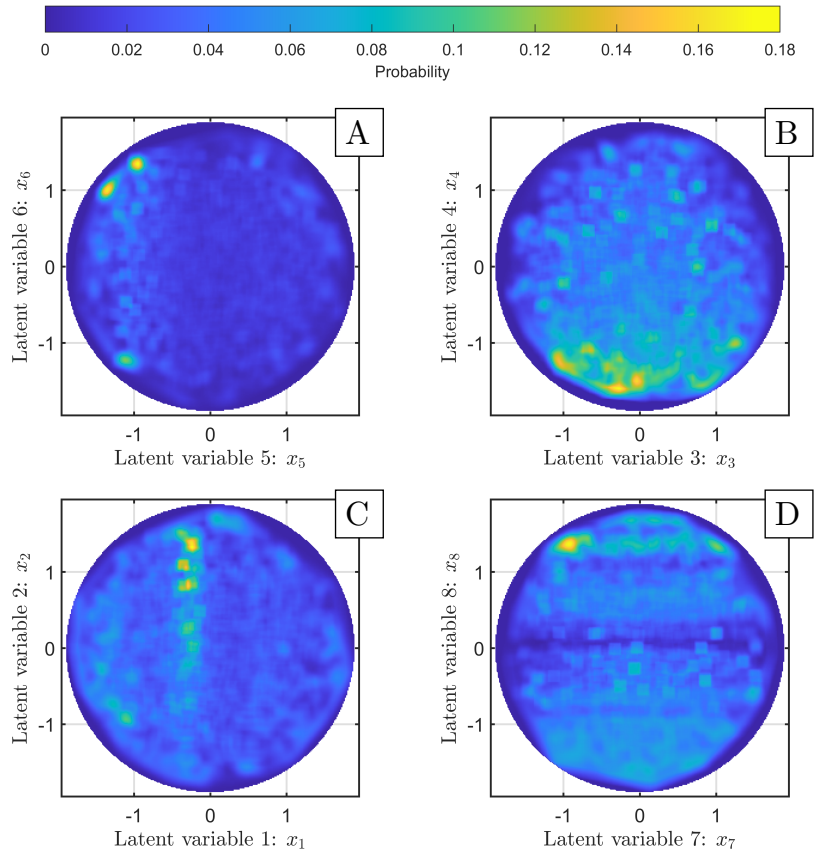

The joint distribution $( P_{\mathbf{x}^*}(\mathbf{x}_{i,\text{can}})$ provides interpretability into the latent space, as it reflects the probability that specific designs are predicted to be globally optimal. For the battery manifolds, we plot four marginal distributions of the eight latent variables in Figure~\ref{fig:latent_results}. These distributions reveal that different latent variables exhibit distinct relationships with design optimality. For example, Figure~\ref{fig:latent_results} shows that the latent variable $x_1 = -0.2$ is likely to produce a high-performing design, approximately Pareto optimal in both charge capacity and average charge voltage. In contrast, $x_2$ displays a broader distribution of values associated with optimal designs, ranging from $ x_2 = -0.6 $ to $ x_2 = 1.5 $. This suggests that $ x_2 $ is associated with a feature that contributes to different performance objectives at different values. Here, we define a feature as a semantically meaningful visual attribute or structural characteristic in the generated design that varies in response to changes in the generative model’s latent variables. By examining the marginal distributions, we can identify which latent variables control features linked to desirable performance outcomes and identify what these features are. 

\section{Concluding Remarks}
In this paper, we have demonstrated how generative adversarial neural networks can be used to design systems with heterogeneous design spaces. The key contributions of this work are twofold: (i) we show that identifying design archetypes enables the formation of a training data set to train generative machine-learning models for when limited or not prior data is available, and (ii) we introduce a systematic approach to enhance the interpretability of latent variables by relating them to both design features and predicted performance. When applied to flow battery manifolds, our approach led to estimated improvements of approximately 3\% in charge capacity and 7\% in average charge voltage compared to the initial set of designs as obtained from computer experiments. These improvements are significant, considering that they were achieved solely by modifying the manifold design.

While our results are promising, several challenges remain that motivate future work. First, our framework has been tested on a single engineering problem. To establish its generalizability, it should be applied to a broader range of engineering systems. Second, the selection of design archetypes in this study was choosing using intuition, yet their diversity likely influences how well the latent space represents the admissible designs. Future work should include an investigation into systematic methods for curating diverse and representative training datasets. Finally, interpreting the relationship between latent variables and optimality currently requires manual inference by a designer. Automating this process by developing methods to measure training set diversity could enable the autonomous identification of design features that drive specific objectives. Despite these open challenges, our study expands the applicability of generative machine-learning models in engineering design to problems for which no prior training data sets are available and contributes to improving their interpretability.

\section*{Acknowledgments}
The authors would like to thank Queen's University Belfast for providing the necessary resources and support for this research. 

\bibliographystyle{asmeconf}  
\bibliography{references}

\end{document}